\documentclass[runningheads]{llncs}
\usepackage{graphicx}
\usepackage{amsmath,amssymb} 
\usepackage{color}
\usepackage{authblk}

\usepackage{subfig}
\usepackage{multirow}
\usepackage{floatrow}
\usepackage{blindtext}
\usepackage{capt-of}
\usepackage{comment}
\usepackage{array,makecell}
\usepackage{amssymb}
\usepackage{pifont}
\newcommand{\cmark}{\ding{51}}%
\newcommand{\xmark}{\ding{55}}%
\usepackage{booktabs}
\usepackage{float}

\floatstyle{plaintop}
\restylefloat{table}
\begin{document}
\pagestyle{headings}
\mainmatter


\title{Finding Correspondences for Optical Flow and Disparity Estimations using a Sub-pixel Convolution-based Encoder-Decoder Network  } 

\titlerunning{ }
\authorrunning{Juan Luis Gonzalez, Muhammad Sarmad}
\author{Juan Luis Gonzalez\footnote[1]{These authors contributed equally}, Muhammad Sarmad*, Hyunjoo J.Lee, Munchurl Kim}
\institute{Korea Advanced Institute of Science and Technology}

\maketitle

\begin{abstract}
Deep convolutional neural networks (DCNN) have recently shown promising results in low-level computer vision problems such as optical flow and disparity estimation, but still, have much room to further improve their performance. In this paper, we propose a novel sub-pixel convolution-based encoder-decoder network for optical flow and disparity estimations, which can extend FlowNetS and DispNet by replacing the deconvolution layers with sup-pixel convolution blocks. By using sub-pixel refinement and estimation on the decoder stages instead of deconvolution, we can significantly improve the estimation accuracy for optical flow and disparity, even with reduced numbers of parameters. We show a supervised end-to-end training of our proposed networks for optical flow and disparity estimations, and an unsupervised end-to-end training for monocular depth and pose estimations. In order to verify the effectiveness of our proposed networks, we perform intensive experiments for (i) optical flow and disparity estimations, and (ii) monocular depth and pose estimations. Throughout the extensive experiments, our proposed networks outperform the baselines such as FlowNetS and DispNet in terms of estimation accuracy and training times.              
\end{abstract}

\section{Introduction}

Vision-based problems have benefited from recent advances in deep learning. The past decade has witnessed a tremendous growth in employment of convolution neural networks (CNN) as a solution to vision-related challenges \cite{cnn1,cnn2}. CNNs have the ability to learn feature maps from the given training data, these learned features can embed complex representations of the input data. The efforts to enhance the performance of CNN have led to various architectures having the skip connections \cite{res1} and deeper layers \cite{cnn1}. 

Recently, encoder-decoder architectures have helped in providing end-to-end learning for many complex vision tasks such as semantic segmentation, optical flow, disparity and scene-flow estimations, etc.\cite{fischer,nikolaus,fcnSemantic2}. The encoder part extracts the feature maps according to the scope of the problem whereas the decoder combines and upscales the desired features to the target resolutions. This architecture, which is combined with the ability of the CNN to learn any kind of end-to-end mappings, has become a very flexible neural-net for solving many vision problems. A common choice for the upscaling and aggregation of features in the decoder part is to use a standard deconvolution operation \cite{fcnSemantic2}. These pipelines have enabled end-to-end learning which is desirable. Nevertheless, deconvolution often results in the so-called checkerboard artifacts. These artifacts appear when the deconvolution layers have "uneven overlap". More precisely speaking, this occurs when the kernel size is not divisible by the stride \cite{odena}. Since the up-scaling of spatial resolution has been intensively studied for super-resolution \cite{srcnn,vdsr,shi}, our idea of using sub-pixel convolution was motivated to replace the deconvolutions in the decoder part. 

So, in this paper, we propose a modified encoder-decoder model by interpreting the transposed convolution (deconvolution) as upscaling operation. Deconvolution has been a standard choice for upscaling in the decoder part ever since it was formally introduced by Long et al.\cite{fcnSemantic}. In principle, we replace the deconvolution operation with an efficient sub-pixel convolution module. This sub-pixel convolution layer has been inspired by the work of Shi et al.\cite{shi}. The benefit of this approach versus other super-resolution networks is that the Efficient Sub-Pixel Convolution Network (ESPCN) operates in the Low Resolution (LR) domain instead of the High Resolution (HR) domain, making it computationally less expensive. Shi et al. \cite{shi2} gave insight that architectures with convolutions purely in the LR domain have more representation power than a model which first up-samples the input images and performs convolution for them afterward. Therefore, in this paper, we not only demonstrate a new encoder-decoder pipeline based on ESPCN but also give a general guideline to potentially convert any model that uses transposed convolution into an ESPCN-based architecture. In order to demonstrate the generality of our idea, we apply it to a wide variety of tasks e.g. optical flow, disparity, and structures from motion.
    
FlowNet\cite{fischer} was devised as an encoder-decoder architecture for optical flow estimation. Optical flow is the apparent motion of objects in a scene. In this paper, our encoder-decoder architecture with sub-pixel convolution for optical flow estimation is named as Flospnet which stands for an optical flow sub-pixel network. In the Flospnet, its encoder part is essentially the same as that of the FlowNet, whereas the decoder part of the Flospnet is different. In the FlowNet, every stage of the decoder part consists of the concatenated output of deconvolution, the skip connection from the encoder and an intermediate prediction layer. The Flospnet maintains the same entire architecture except for replacing the deconvolution layers with sup-pixel convolution modules. 

The disparity estimation baseline network, DispNet\cite{nikolaus}, can be changed in a similar way as done for the Flospnet. In this case, our encoder-decoder pipeline architecture with sub-pixel convolution for disparity estimation is called Despnet which stands for a disparity efficient sup-pixel network. The difference between the Flospnet and the Despnet is the numbers of output channels at each sup-pixel convolution module in the decoder parts. All ESPCN modules in the Flospnet yield two output channels since it estimates optical flows in horizontal and vertical directions while each sup-pixel convolution module in the Despnet has only one output channel because the disparity is estimated only along the horizontal direction.       

As mentioned before, our Despnet is devised for unsupervised monocular depth estimation by replacing the deconvolution of the original DispNet by Zhou et al.\cite{ZhouBSL17} with the sub-pixel convolution. We also adopt the sub-pixel convolution for the decoder part of the pose network in their work, thereby completely eliminating any deconvolution layer.

In our experiments, we show that the Flospnet and Despnet significantly outperform their corresponding baseline architectures \cite{fischer,nikolaus,ZhouBSL17}. The superiorities of the Flospnet and Despnet come from the following two reasons: (i) the sub-pixel convolution performs more precisely the mapping from a lower dimension to higher dimensions than the deconvolution operation where it is often adopted in super-resolution problems; and (ii) the both models are fast trained since our sub-pixel convolution models use a fewer number of parameters.                                              
Consequently, any network pipeline containing deconvolution layers can be replaced by sub-pixel convolution, which can lead to improved performance and a less number of network parameters. We can summarize the contributions of our work as follows:
    
1) {\it Flospnet}: It is designed based on the FlowNet by adopting sub-pixel convolution for optical flow estimation, which has brought a significant improvement on the estimation accuracy. 
 
2) {\it Despnet}: It is a novel disparity estimation network after replacing the standard deconvolution layers in DispNet with sub-pixel convolution. We also show a variation of Despnet with rectangular convolution which outperforms all trained models.

\section{Related Works}

\subsection{Optical Flow} Optical flow has been investigated by various techniques from the era of Horn and Shunck \cite{Horn&Shunck}. Such techniques have been combined in various ways to deal with large displacements and combinational correspondence, trying to improve estimation accuracy \cite{LargeDispopticalFlow,LargeDispOpticalFlowDeep,epicflow}. However, these techniques are not learning-based methods that can be trained to effectively solve the difficult optical flow estimation problems for complex moving sequences.

Recently, deep learning-based optical flow estimation has drawn much attention and has shown remarking performance improvement \cite{opticalFlowwithGeometricOcclusion,6751324,NIPS2013_5193}. Fischer et al. \cite{fischer} have used a CNN-based optical flow estimation network, called the FlowNet. The FlowNet adopts an encoder-decoder architecture by which the features for optical flows between two consecutive input frames are extracted in its encoder part and the estimated optical flow field is constructed in the decoder part through multi-layer deconvoltion operations. The FlowNet shows a limited performance by yielding artifacts in estimated optical flows along moving object boundaries. SpyNet\cite{spynet} uses a fraction of parameters compared to the FlowNet but it is not end-to-end trainable. Ilg et al.\cite{flownet2} have devised the FlowNet2, which is a large-sized network, by combining five FlowNets as basic building blocks in a cascaded and parallel manner to improve the estimation accuracy. LiteFlowNet\cite{hui18liteflownet} uses fewer parameters than the FlowNet2 but shows better or comparable performance depending on test datasets.  

\subsection{Disparity and Structure from Motion}
Disparity estimation has been extensively studied using deep learning based methods\cite{semiglobal,ZhangStereo,ZbontarL15}. 
None of these methods provided an end-to-end learning architecture. Then, Mayer et al.\cite{nikolaus} adopted the FlowNet architecture as an end-to-end disparity estimation pipeline, called the DispNet. They added one up-scaling stage and an additional intermediate convolution layer into the decoder part of the FlowNet for the DispNet. Structure from motion (SFM) is a technique of estimating 3D structures from 2D image sequences and camera pose/motion signals. SFM has been recently investigated in an unsupervised fashion by a number of works. Liu et al. \cite{LiuSLR15} devise a single image deep convolutional neural field (DCNF) model for unsupervised depth estimations by exploring conditional random fields and super-pixel pooling. In contrast, for stereo pairs, Garg et al. \cite{GargBR16} trained an encoder-decoder architecture for unsupervised monocular depth estimation by synthesizing a backward-warped image using the estimated left disparity map and the right image. The warped left images are used to calculate the reconstruction error. Similarly, Godard et al. \cite{GargBR16} used not only the right consistency but also the left-right consistencies to train an unsupervised disparity estimation network, thus greatly improving the estimation performance. In a different approach, Zhou et al.\cite{ZhouBSL17} exploited the relative pose information in a sequence of images to train a disparity network and a pose estimation network in an unsupervised manner by performing view synthesis of the center frame (source image) to the different reference frames (target images).

\subsection{Deconvolution and Super Resolution}
The so-called deconvolution layer finds its roots back to the work of Long et al. \cite{fcnSemantic}, where it was introduced for semantic segmentation for the first time as a way of up-sampling using backward strided convolution. Noh et al.\cite{fcnSemantic2} extended it by using a deeper decoder part with many stages of deconvolution layers for semantic segmentation. Since then, it has been adopted widely by many works (e.g. \cite{nikolaus}\cite{sfmlearner}\cite{fischer}) due to its ease of use and end-to-end training capability. Super-resolution has been a popular research topic within the video processing and photo enhancement community. SRCNN \cite{srcnn} first showed the applicability of CNNs for super-resolution problems and later VDSR\cite{vdsr} significantly improved its performance by incorporating a much deeper architecture with residual learning. However, these techniques require expensive convolution operations in the HR domain. In contrast, ESPCN \cite{shi} defines a network pipeline where all operations are carried out in the LR domain. Upscaling in ESPCN is achieved by rearranging the pixels of the last layer's \(r^2\) feature maps where \(r\) is the upscaling ratio.

\section{Method}
In this section, we will highlight in details our encoder-decoder architecture with sub-pixel convolution for optical flow, depth, and SFM estimations. For this, we will shed light on our modified pipelines for optical flow, disparity, and SFM, respectively. At the same time, we will compare our modified pipelines with their corresponding baseline networks in the previous works \cite{nikolaus,fischer,ZhouBSL17}.
\subsection{Despnet}

\begin{figure}
	\centering
	\includegraphics[scale=0.35]{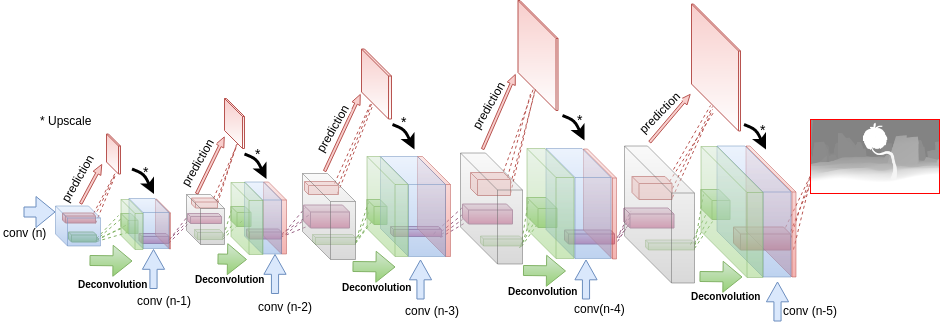}
	\caption{Decoder of DispNet\cite{nikolaus}. {\it Green Volume}: Deconvolution layers. {\it Blue Volume}: Skip connections from Encoder. {\it Red Volume} : Predictions and Upscaled Predictions. {\it Gray Volume} : Convolution layer for concatenated volumes}
	\label{fig:dispnet}
\end{figure}

Despnet is a natural extension to the DispNet\cite{nikolaus}. Fig.~\ref{fig:dispnet} shows the decoder part of the DispNet that contains skip connections from the encoder part (blocks in light blue color). The light green blocks indicate the deconvolution layer outputs. The prediction operation is a standard convolution layer which outputs a single channel, represented by the red boxes on top of the convolution layers in Fig.~\ref{fig:dispnet}. This prediction output is then upscaled and concatenated with the deconvolutions (boxes in light green color) and the skip connections (boxes in light blue color) as indicated by the black arrows in  Fig.~\ref{fig:dispnet}. The concatenated channels are then convolved once more to form the "iconv" output which is shown as the boxes in light gray color. 

\begin{figure}
	\centering
	\includegraphics[scale=0.33]{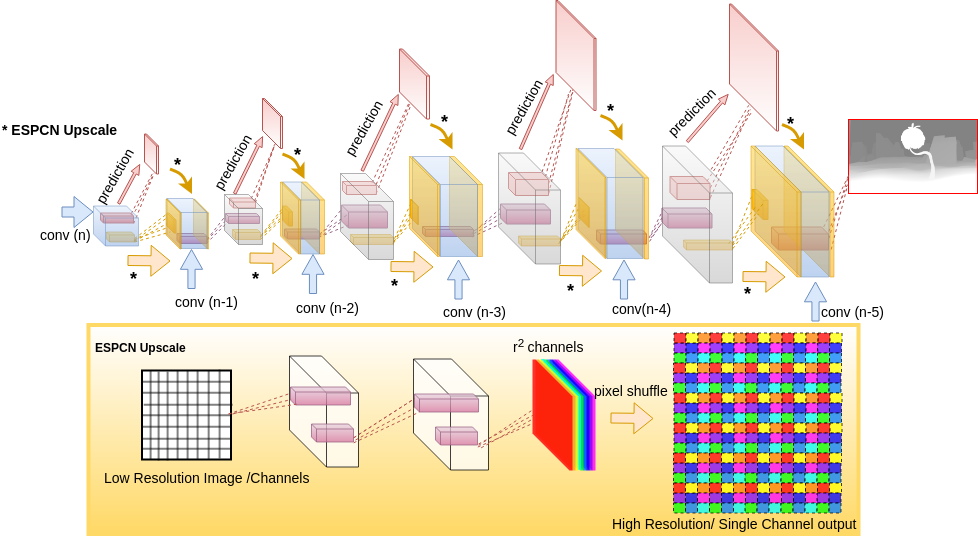}
    \caption{Decoder part of Despnet. {\it Yellow Volume}: Sup-pixel convolution layers. {\it Blue Volume}: Convolution Skip connections from Encoder. {\it Red Volume} : Predictions. {\it Gray Volume} : Convolution layer for concatenated volumes. Our Despnet replaces all Deconvolution operations and Upscale Operations by sup-pixel convolution modules, hence the replaced volume shown in yellow has only a single channel. The sup-pixel convolution block is composed by 2 conv-relu layers and a conv-pixel shuffle operation as in \cite{shi}. For all our experiments the upscaling factor "r" is set to 2 }
	\label{fig:despnet}
\end{figure}

As mentioned earlier, the deconvolution layers can be substituted by the sub-pixel convolution modules which have a more powerful mapping capability from lower to higher dimensions of feature maps. So, we extend the DispNet to a new pipeline with sub-pixel convolution which is called the Despnet. Fig.~\ref{fig:despnet} shows the architecture of our extended Despnet. In Fig.~\ref{fig:despnet} all the deconvolution and upscaling operations have been replaced by sub-pixel convolution blocks. The yellow volumes in Fig.~\ref{fig:despnet} indicate the sub-pixel convolution module, which is highlighted in the bottom yellow box of Fig.~\ref{fig:despnet}. It is important to note here that the input to this sub-pixel convolution module can be single or multiple channels but the sub-pixel convolution module always produces a single channel output (or two channels for the Flospnet case). This can be noted as the thicknesses of the yellow volumes in Fig.~\ref{fig:despnet} which is narrower than those of the green volumes. It should be noted in Fig.~\ref{fig:despnet} that the resulting Despnet has far fewer parameters compared to the DispNet. A detailed comparison between the DispNet vs the Despnet is shown in Table~\ref{table:dispANDflow} of the Experimental Results section.


\subsection{Flospnet}
\begin{figure}
	\centering
	\includegraphics[scale=0.35]{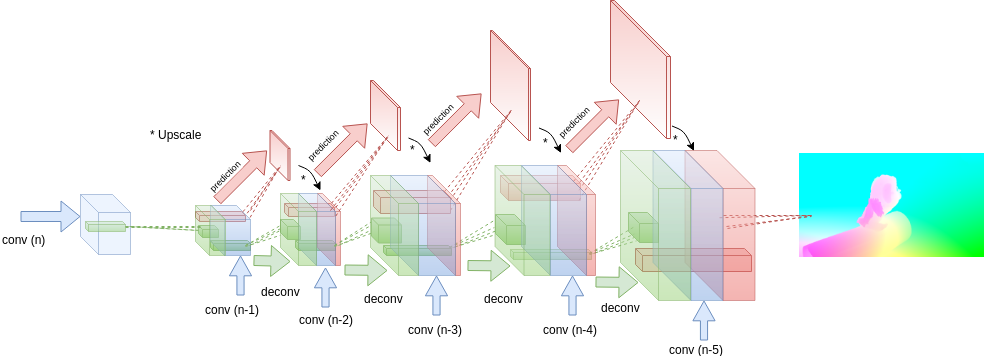}
	\caption{Decoder of FlowNet\cite{fischer}. {\it Green Volume}: Deconvolution layers. {\it Blue Volume}: Convolution Skip connections from Encoder. {\it Red Volume}: Predictions and Upscaled Image} 
	\label{fig:flownet}
\end{figure}

Flospnet is a modified FlowNet\cite{fischer} where the deconvolution operation is replaced with sub-pixel convolution. That is, its decoder pipeline has used sub-pixel convolution modules in place of the deconvolution layers. Fig.~\ref{fig:flownet} shows the standard decoder architecture. The decoder part of the Flospnet is formed from the FlowNet in the same way as the Despnet was formed from the DispNet as shown in Fig.~\ref{fig:despnet}. The encoder part of the Flospnet is similar to that of the FlowNet. Again, due to use of the sub-pixel convolution modules, our Flospnet has a smaller number of parameters compared to the FlowNet. A detailed comparison of their performance and numbers of parameters is shown in Table~\ref{table:dispANDflow} of the Experimental Results section. 

\subsection{An Unsupervised SFM Network with Sub-pixel Convolution}
In order to demonstrate the effectiveness of the sub-pixel convolution modules for a practical problem such as 'structure from motion (SFM)', we adopt the Despnet instead of DispNet \cite{nikolaus} for disparity estimation in unsupervised SFM learning by Zhou et al.\cite{ZhouBSL17}. In addition, we further replace their PoseNet with a sub-pixel convolution based PoseNet. The resulting architecture can be formed in a similar way of forming the Despnet, where training is computationally less expensive due to the usage of a smaller number of parameters. Details of performance for unsupervised monocular depth estimation can be found in Table~\ref{table:sfm} of Experimental Results section. 

\section{Experimental Results}
\subsection{Dataset}
\subsection*{Monkaa and Driving}
Mayer et al. \cite{nikolaus} have provided an open source synthetic dataset which contains comprehensive ground truths for training optical flow and disparity. We train and test our networks on two out of three of their datasets i.e. Monkaa and Driving.

\begin{figure}
	\centering
	\includegraphics[scale=0.64]{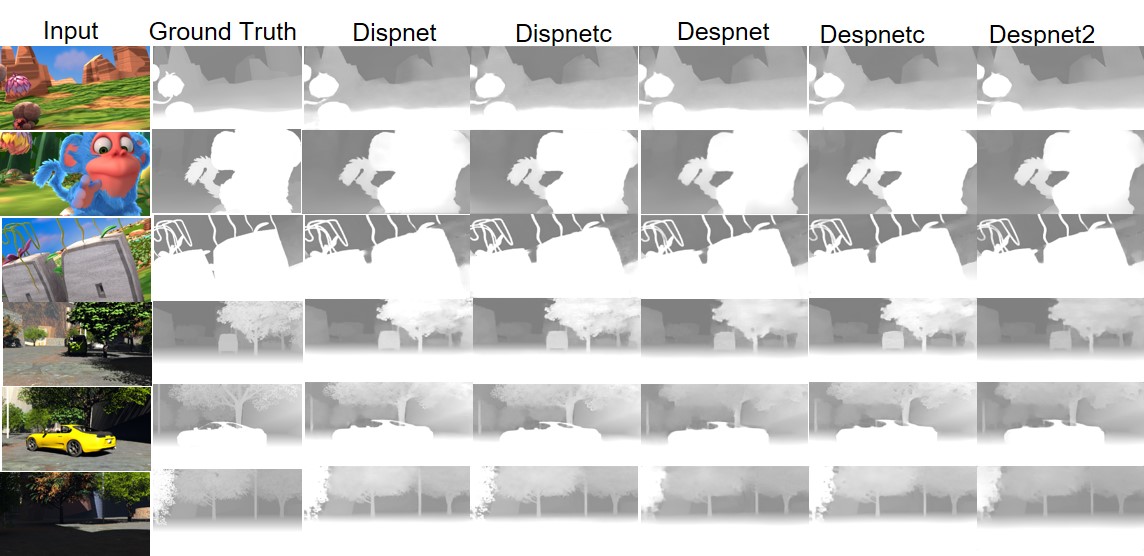}
    \caption{Disparity Results for various Networks along with Input Image and Ground Truth}
	\label{fig:Disp1}
\end{figure}

\begin{figure}
	\centering
	\includegraphics[scale=1.0]{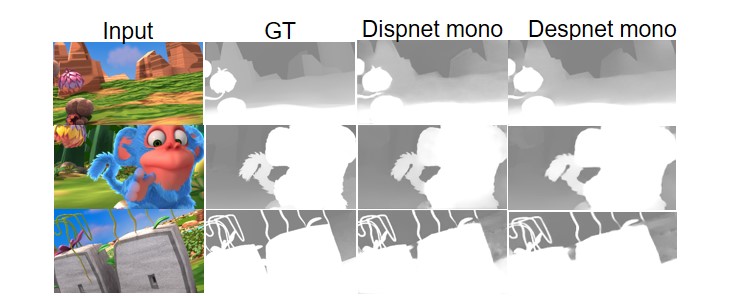}
    \caption{Disparity Results for Monocular Networks along with Input Image and Ground Truth}
	\label{fig:Disp2}
\end{figure}

\begin{figure}
	\centering
	\includegraphics[scale=0.6]{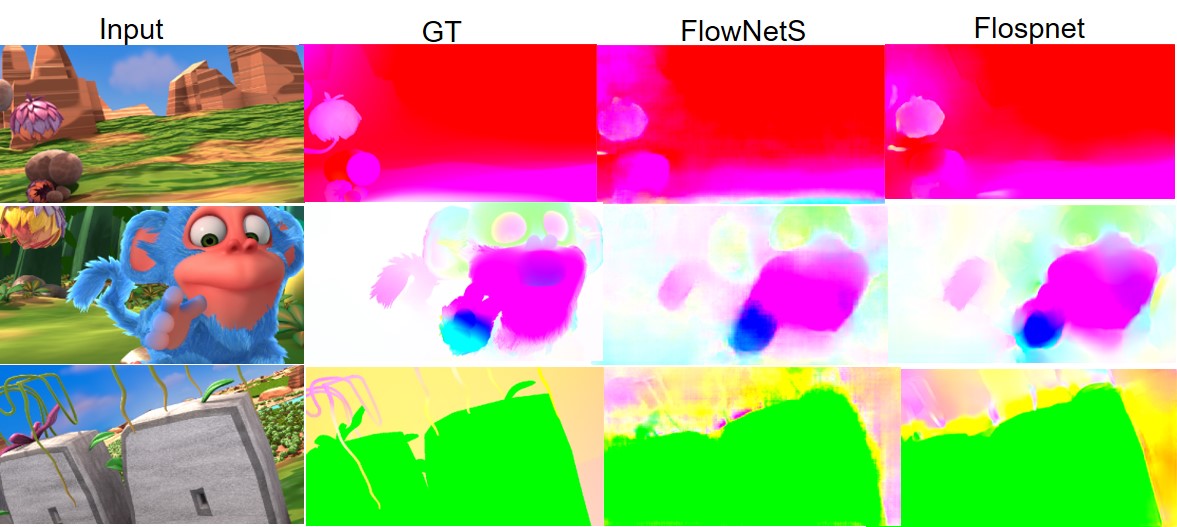}
    \caption{Optical Flow Results for various Networks along with Input Image and Ground Truth. Our Flospnet provides better optical flow than the baseline}
	\label{fig:Flow1}
\end{figure}

For Disparity and optical flow estimations, the results for both datasets are shown in Table~\ref{table:dispANDflow}. All shown errors are end point errors (EPE), and additionally, the number of parameters used in each network is also tabulated.
The Despnet2 is a modified Despnet with rectangular kernels i.e. 3x7, 3x5, 3x5 in the first three layers of the encoder. Whereas, the Despnet-mono is a Despnet trained on a single image instead of stereo image pairs. Fig.~\ref{fig:Disp1} and Fig.~\ref{fig:Disp2} show the output for our trained Despnet-based pipeline and the standard DispNet\cite{nikolaus}. Similarly, Fig.~\ref{fig:Flow1} shows and compares the output optical flow for the Flospnet and FlowNet \cite{fischer} networks.

\setlength{\tabcolsep}{4pt}
\begin{table}
\begin{center}
\caption{
Network Performance Metrics: All error metrics for both data sets are in EPE (End Point Error). Inference time is in seconds (s), tested on a GTX 1080 GPU.}
\label{table:dispANDflow}
\begin{tabular}{p{3cm}|>{\centering\arraybackslash}p{1.5cm}>{\centering\arraybackslash}p{1.5cm}>{\centering\arraybackslash}p{1.5cm}>{\centering\arraybackslash}p{2cm}}
\hline\noalign{\smallskip}
Method $\qquad\qquad$&  Params (Mils)  & Monkaa (EPE) & Driving (EPE) & Inference time\\
\noalign{\smallskip}
\hline
\noalign{\smallskip}
DispNet       & 42  & 5.314 & 10.927 & 0.003\\
Despnet       & 31  & 3.914 & 8.785  & 0.004\\
Despnet2      & 30  & \bf3.715 & \bf5.467 & 0.004\\
DispNetC      & 42  & 4.314 & 10.885 & 0.008\\
DespnetC      & 32  & 4.054 & 10.803 & 0.009\\
DispNet-mono   & 42  & 4.345 & - & 0.003\\
Despnet-mono   & 31  & \bf4.118 & - & 0.004\\
FlowNet       & 42  & 6.860 & - & 0.003\\
Flospnet      & 31  & \bf3.684 & - & 0.005\\
\hline
\end{tabular}
\end{center}
\end{table}

\subsection*{KITTI}
\begin{figure}
	\centering
	\includegraphics[scale=0.6]{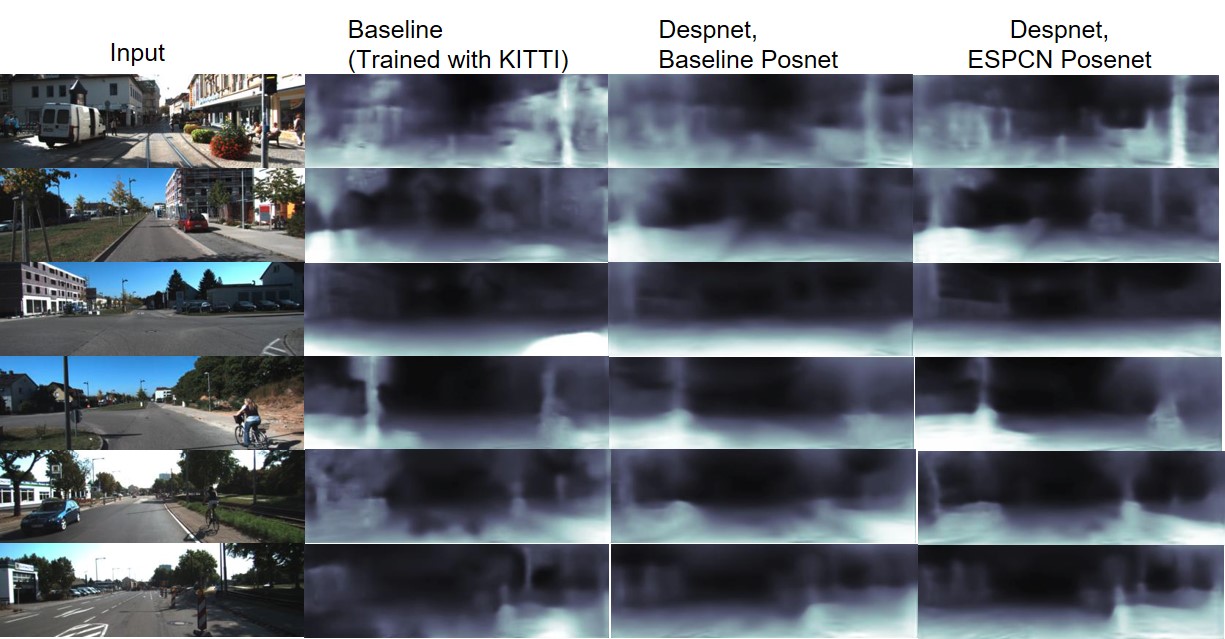}
    \caption{Disparity Results for various monocular unsupervised trained networks}
	\label{fig:sfm1}
\end{figure}

We used the KITTI dataset to train our modified unsupervised SFM network with sub-pixel convolution. Essentially, we changed the DispNet and PoseNet architectures in the implementation of Zhou et al.\cite{ZhouBSL17} with sub-pixel convolution modules. This change in architecture was implemented in a similar way as the case of the Despnet from the DispNet\cite{nikolaus}. We used the implementation of their unsupervised SFM network as the baseline. Table~\ref{table:sfm} compares our modified SFM network and some previous networks. Fig.~\ref{fig:sfm1} shows the output comparison of the baseline (trained on KITTI only) and our modified SFM network.

\setlength{\tabcolsep}{0.1pt}
\begin{table}
\begin{center}
\caption{
Monocular depth results for KITTI dataset: Note that  K = KITTI, and CS = Cityscapes. 
}

\label{table:sfm}
\begin{tabular}{@{\extracolsep{2.8pt}}p{2.0cm}>{\centering\arraybackslash}p{1.4cm}>{\centering\arraybackslash}p{0.7cm}>{\centering\arraybackslash}p{0.7cm}>{\centering\arraybackslash}p{0.9cm}>{\centering\arraybackslash}p{0.9cm}>{\centering\arraybackslash}p{0.9cm}>{\centering\arraybackslash}p{0.9cm}>{\centering\arraybackslash}p{1.0cm}>{\centering\arraybackslash}p{1.0cm}>{\centering\arraybackslash}p{1.0cm}@{}}

\hline\noalign{\smallskip}

Method & Dataset  & \multicolumn{2}{>{\centering\arraybackslash}p{1.5cm}}{ESPCN Module} & \multicolumn{4}{c}{Error}   &  \multicolumn{3}{c}{Accuracy} \\\cline{3-4}\cline{5-8}\cline{9-11}
\noalign{\smallskip}

\noalign{\smallskip}
&  & {\scriptsize\bf Disp Net} & {\scriptsize\bf Pose Net} &  {\tiny\bf Abs Rel} & {\tiny\bf Sq Rel} & {\tiny\bf RMSE} & {\tiny\bf RMSE log} & \tiny\bf${\it \delta<1.25}$ & \tiny\bf${\it\delta<1.25^2}$ &\tiny\bf${\it \delta<1.25^3}$  \\
\noalign{\smallskip}
\hline 
\noalign{\smallskip}
Zhou et al.\cite{ZhouBSL17} & K      & - & - & \scriptsize0.208 & \scriptsize1.768 & \scriptsize6.856 & \scriptsize0.283 & \scriptsize0.678 & \scriptsize0.885 & \scriptsize0.957\\
Zhou et al.\cite{ZhouBSL17} & CS     & - & - & \scriptsize0.267 & \scriptsize2.686 & \scriptsize7.580 & \scriptsize0.334 & \scriptsize0.577 & \scriptsize0.840 & \scriptsize0.937\\
Zhou et al.\cite{ZhouBSL17} & K + CS & - & - & \scriptsize0.198 & \scriptsize1.836 & \scriptsize6.565 & \scriptsize0.275 & \scriptsize0.718 & \scriptsize0.901 & \scriptsize0.960\\
\bf Ours                        & K      & \cmark & \xmark& \scriptsize\bf0.189 & \scriptsize\bf1.446 & \scriptsize6.845 & \scriptsize0.277 & \scriptsize0.705 & \scriptsize0.891 & \scriptsize0.958\\
\bf Ours                        & K      & \cmark & \cmark & \scriptsize0.190 & \scriptsize1.559 & \scriptsize\bf6.267 & \scriptsize\bf0.264 & \scriptsize\bf0.719 & \scriptsize\bf0.906 & \scriptsize\bf0.965\\
\hline


\end{tabular}
\end{center}
\end{table}

\subsection{Implementation Details}
Our architectures (Flospnet, Despnet and SFM network) are implemented in Pytorch \cite{torch} for all experiments. Our implementations for disparity and optical flow are made on top of the implementation of Pinnard\footnote[1]{https://github.com/ClementPinard/FlowNetPytorch}. For the evaluation of our SFM network, we build it on the implementation of Pinnard\footnote[2]{https://github.com/ClementPinard/SfmLearner-Pytorch}. 

To perform the experiments for optical flow estimation networks (FlowNet etc.), we use on-the-fly data augmentation with random modifications of crop, rotate, translate, horizontal flip and vertical flip. Whereas, for stereo disparity estimation, we use the same except for random translate and rotate since these transformations would break the epipolar constraint \cite{nikolaus}.

For training our models, the batch size was set to eight (images, stereo pairs, sequences, etc.). We employed a learning rate of 0.0001, the number of epochs were 300, and the learning-rate-decay-schedule was set to [100,150,200] with a decay of 0.5. We used Adam optimizer for all our experiments. We used the same losses as used in the baselines for training all our networks.

\section{Discussions}
In this section, we discuss the implications of using the sub-pixel convolution of the ESPCN\cite{shi} for our extended networks for optical flow, disparity, and SFM estimations. 

The intuition behind our extended networks is that by incorporating a better up-sampling mechanism free of checkerboard artifacts like the ones due to deconvolution, our extended networks are able to learn finner LR to HR mapping in feature map generation throughout the decoder stages. In addition, by using the sub-pixel convolution modules, we can increase the depths of the networks and can hence enlarge the receptive fields in a very effective way as shown in Fig.~\ref{fig:despnet}. So we come to reduce the number of parameters compared to the baselines. Each sub-pixel convolution module adds up to three layers of convolution, making a total addition of thirty layers to the networks (Despnet, Flospnet, etc.). Each sub-pixel convolution module sandwiches the skip connections from the encoder stages, which can be viewed as a spatial attention mask. It is also pertinent to note in Fig.~\ref{fig:despnet} that the skip connections are a vital part of this pipeline since the attention on its own is not useful without structural representations that are provided by the skip connections. 

\subsection{Flospnet vs. FlowNet}
The Flospnet improves the FlowNet\cite{fischer} in two aspects: Not only were we able to reduce the number of parameters required by the network but also the EPE is reduced, as shown in Table~\ref{table:dispANDflow}. It can be seen in Fig.~\ref{fig:Flow1} that our Flospnet produces crisper optical flow than the baseline (FlowNet). We approximately achieved a 31\% improvement in EPE over the baseline whereas an 18\% reduction in the number of parameters was obtained, as shown in Table~\ref{table:dispANDflow}. From the observation of these remarkable improvements, other optical flow frameworks can also benefit from using the sub-pixel convolution instead of the deconvolution operation. 

\subsection{Despnet vs. DispNet}
The Despnet also greatly improves the DispNet in terms of the number of parameters used by the network and the EPE values, as shown in Table~\ref{table:dispANDflow}. Apart from this, we also trained and tested the Despnet2 which benefits from the rectangular kernels as explained in the Experimental Results section. The usage of such rectangular kernels makes a natural sense in disparity estimation for stereo images where the correspondences are often found in the horizontal axis. Therefore using a wider kernel in horizontal direction helps finding precise disparity while using a shorter kernel in vertical reduces the number of parameters required in usage. We have also tested a DispNetC\cite{nikolaus} which uses a 1D correlation layer. After applying the sub-pixel convolution to the decoder part of the DispNetC, we get the DespnetC which performs better than the DispNetC. We have used a max displacement of 35 in the 1D correlation layer for both the DispNetC and the DespnetC. For more information on correlation layer, the reader can refer to Mayer et al.\cite{nikolaus}. We noticed that the DispNet-mono, which is trained on single images, performed better than the DispNet, which is trained on stereo image pairs. The Despnet-mono is also extended by applying the sub-pixel convolution to the decoder part of the DispNet-mono, which performs even better in terms of EPE. Fig.~\ref{fig:Disp1} shows the output of the Despnet in comparison with its baseline (Dispnet). Fig.~\ref{fig:Disp2} compares the outputs of the Despnet-mono and the Dispnet-mono.

\begin{figure}
	\includegraphics[scale=0.7]{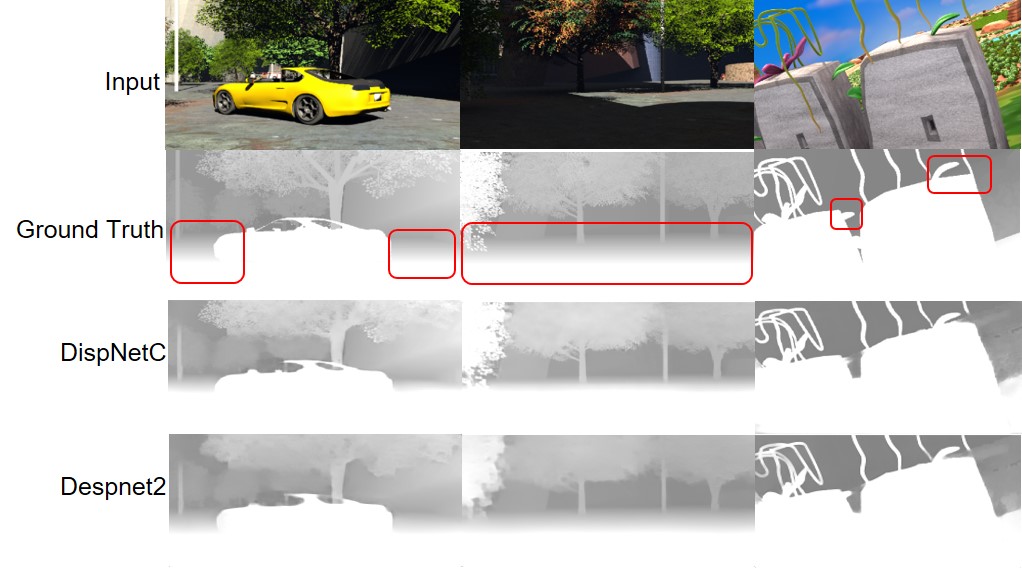}
	\caption{Comparison of best disparity estimation networks with and without sup-pixel convolution modules. Despnet2 achieves the best disparity estimation. Key areas of comparison are highlighted by red boxes}
	\label{fig:disc1}
\end{figure}

As shown in Fig.~\ref{fig:Disp1}, it can be seen that the Despnet approaches can produce crisper disparity estimation for background objects and more accurate estimation for the objects close to the cameras. This can be easily observed in the highlighted areas in Fig.~\ref{fig:disc1}, where the DispNetC fails to estimate the vanishing gradients in the disparity of the asphalt for the picture in the first column, while our Despnet accurately reproduces the disparity map. Similarly, other highlighted areas in the right columns show the superior performance of the sub-pixel convolution based architecture. The experimental results in Table~\ref{table:dispANDflow} clearly demonstrate the smaller EPE values achieved by the sub-pixel convolution based architectures for different datasets. For instance, in the case of the Despnet, we achieved a 26\% reduction in both the EPE and the number of parameters for Monkaa dataset. Furthermore, the Depsnet2 attains an even lower EPE and smaller numbers of parameters than the Despnet due to the usage of the rectangular kernels.   

\subsection{Despnet-based SFM network vs. DispNet-based SFM network}
In addition to applying the Despnet to the unsupervised SFM network, we also changed the PoseNet in the implementation of Zhou et al.\cite{ZhouBSL17}. As an ablation study, we used only the Despnet while keeping the PoseNet as is in the baseline. Table~\ref{fig:sfm1} shows the SFM estimation performance for ours and the baseline (DispNet-based SFM network). It can be observed in Table~\ref{fig:sfm1} that an incremental improvement is seen over the baseline using the sub-pixel convolution. A more interesting aspect is that replacing the PoseNet with the sub-pixel convolution based network further improves the SFM estimation performance in terms of RMSE, such that our network trained on KITTI alone outperforms the baseline trained on both KITTI and CityScapes. 

As shown in Fig.~\ref{fig:sfm1}, the disparity estimation by the Despnet-based SFM network is better than that of the DispNet-based SFM network in perspectives of structural details and disparity smoothness. It can be observed that the baseline disparity estimation for the third row input picture fails by predicting close-by blobs on the bottom right part of the scene which are absent from the image, while the Despnet-based SFM network smoothly generates the disparity map with little artifacts. It can also be seen in the fourth row of Fig.~\ref{fig:sfm1} that the baseline is not able to accurately predict the disparity map of the road ahead, predicting instead very far away objects in the lower left region of the image. Again, our approach prevails in such scenario as well. Additionally, our model can also deal with cluttered and crowded scenes as can be observed in the first row of Fig.~\ref{fig:sfm1} where the baseline performs poorly in discriminating various object's disparities. 

\section{Conclusion}
In this work, we have demonstrated that the sub-pixel convolution can effectively replace the deconvolution operations in the encoder-decoder architectures to solve low-level computer vision problems such as optical flow, disparity, and SFM estimations. The sub-pixel convolution block brings in the merits of improved accuracy, reduced numbers of parameters, and faster training speed. We showed the effectiveness of our extensions to the FlowNet, DispNet and SFM network was consistently achieved with the involvement of sub-pixel convolution from various and extensive experiments. It is also worthwhile to experiment with our idea for other vision tasks such as semantic segmentation, style transfer, and object detection, etc., which we will do. 

\bibliographystyle{splncs}
\bibliography{egbib}

\end{document}